\documentclass[letterpaper, 10pt, conference]{ieeeconf}      

\IEEEoverridecommandlockouts                              

\usepackage{moreverb,url}
\usepackage{endnotes}
\usepackage[acronym,nomain]{glossaries}
\usepackage[dvipsnames, table]{xcolor}
\usepackage{pgf}
\usepackage{pgfplots}
\pgfplotsset{compat=1.14}
\usepackage{float}
\usepackage{booktabs}
\usepackage[detect-none,binary-units=true]{siunitx}
\DeclareSIUnit{\nothing}{\relax}
\usepackage{multirow}
\usepackage{nicefrac}
\usepackage[hidelinks]{hyperref}

\definecolor{somegray}{rgb}{0.5, 0.5, 0.5}
\newcommand{\darkgrayed}[1]{\textcolor{somegray}{#1}}
\makeatletter
\newcommand*\titleheader[1]{\gdef\@titleheader{#1}}
\AtBeginDocument{%
  \let\st@red@title\@title
  \def\@title{%
    \vskip-2.0em
    \bgroup\normalfont\large\centering\@titleheader\par\egroup
    \vskip0.0em\st@red@title}
}

\makeatother

\titleheader{\darkgrayed{This paper has been accepted for publication in the IEEE ICRA 2024 conference\\\copyright 2024 IEEE.}}

\DeclareSIUnit{\nothing}{\relax}
\DeclareSIUnit\mac{MAC}

\title{\LARGE \bf
High-throughput Visual Nano-drone to Nano-drone Relative Localization using Onboard Fully Convolutional Networks}

\author{Luca Crupi$^{1}$, Alessandro Giusti$^{1}$, and Daniele Palossi$^{12}$%
\thanks{This work was partially supported by the Secure Systems Research Center (SSRC) of the UAE Technology Innovation Institute (TII) and the Swiss National Science Foundation (SNSF) through the NCCR Robotics.}%
\thanks{$^{1}$L. Crupi, A. Giusti, and D. Palossi are with the Dalle Molle Institute for Artificial Intelligence (IDSIA), USI-SUPSI, Lugano, 6962, Switzerland
        {\tt\small name.surname@idsia.ch}}%
\thanks{$^{2}$D. Palossi is also with the Integrated Systems Laboratory (IIS), ETH Z\"urich, Z\"urich, 8092, Switzerland}%
}

\begin{document}


\maketitle

\begin{abstract}
Relative drone-to-drone localization is a fundamental building block for any swarm operations.
We address this task in the context of miniaturized nano-drones, i.e., $\sim$\SI{10}{\centi\meter} in diameter, which show an ever-growing interest due to novel use cases enabled by their reduced form factor.
The price for their versatility comes with limited onboard resources, i.e., sensors, processing units, and memory, which limits the complexity of the onboard algorithms.
A traditional solution to overcome these limitations is represented by lightweight deep learning models directly deployed aboard nano-drones.
This work tackles the challenging relative pose estimation between nano-drones using only a gray-scale low-resolution camera and an ultra-low-power System-on-Chip (SoC) hosted onboard.
We present a vertically integrated system based on a novel vision-based fully convolutional neural network (FCNN), which runs at \SI{39}{\hertz} within \SI{101}{\milli\watt} onboard a Crazyflie nano-drone extended with the GWT GAP8 SoC.
We compare our FCNN against three State-of-the-Art (SoA) systems.
Considering the best-performing SoA approach, our model results in a $R^2$ improvement from 32 to 47\% on the horizontal image coordinate and from 18 to 55\% on the vertical image coordinate, on a real-world dataset of $\sim$\SI{30}{\kilo\nothing} images.
Finally, our in-field tests show a reduction of the average tracking error of 37\% compared to a previous SoA work and an endurance performance up to the entire battery lifetime of \SI{4}{\minute}.
\end{abstract}

\section*{Supplementary video material}
In-field tests: \href{https://youtu.be/wMFYnv8UE80}{https://youtu.be/wMFYnv8UE80}.

\section{Introduction} \label{sec:intro}

\begin{figure}[t]
  \includegraphics[width=\columnwidth]{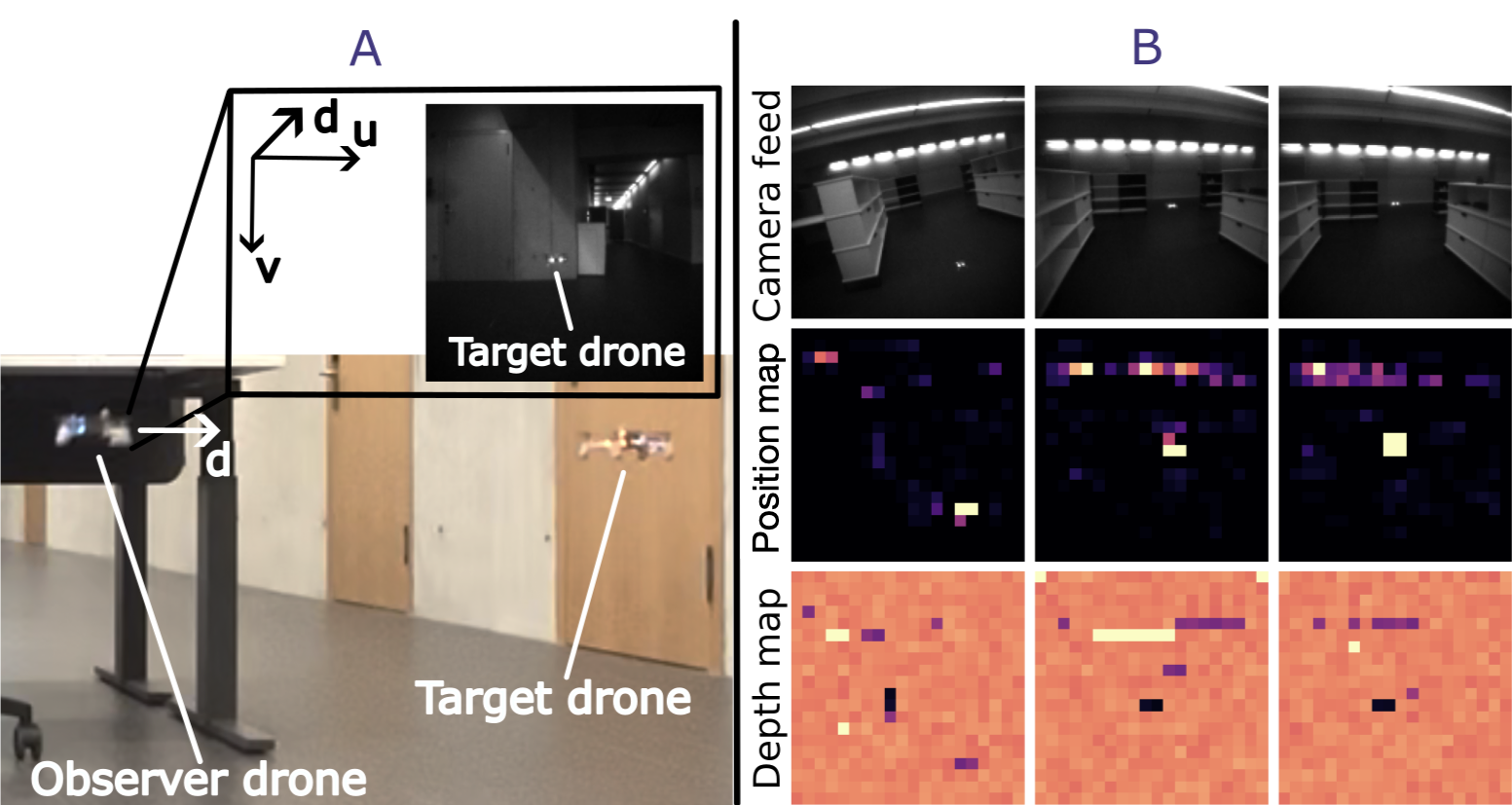}
  \caption{A) The observer nano-drone tracks a target one. B) Three image samples from the onboard camera associated with the position and the depth map computed by the fully convolutional neural network.}
  \label{fig:intro}
\end{figure}

Precise drone-to-drone relative pose estimation is a fundamental skill for many swarm operations~\cite{10150322_Formation, cardona2019robot}.
Drones capable of precisely localizing peers in the flock can adjust their attitude to maintain desired formations~\cite{10150322_Formation} or to optimize their trajectories to maximize their effectiveness~\cite{cardona2019robot}, e.g., in search-and-rescue or inspection missions.
Palm-sized quadrotors, also called nano-drones, represent an uprising class of flying robotic platforms with a weight of less than \SI{50}{\gram} and a sub-\SI{10}{\centi\meter} diameter~\cite{BonatoD2D, LiD2D, moldagalieva2023virtual}, see Figure~\ref{fig:intro}-A.
Thanks to their small form factor, these miniaturized flying robots enable novel application scenarios in cluttered and narrow indoor environments~\cite{DBLP:journals/corr/abs-2304-02359, s19030478_Burgues}, e.g., industrial plants, collapsed buildings, etc., as well as in human surroundings, being harmless even in case of a crash~\cite{frontnet}.
Additionally, nano-drones are extremely cheap platforms compared to traditional \SI{}{\kilo\gram}-scale multi-rotors due to their simplified design and electronics.
Conversely, their size severely constrains onboard resources, such as sensors, memory capacity, and computational power.
For this reason, many autonomous nano-drones run onboard convolutional neural networks (CNNs) for their perception~\cite{BonatoD2D,LiD2D,moldagalieva2023virtual} instead of complex geometrical computer vision pipelines~\cite{palossi2013gpu}.

This work addresses the relative drone-to-drone pose estimation with nano-drones relying only on their onboard hardware.
Among many successful technologies employed in drone pose estimation, some are prevented aboard nano-drones due to power consumption, weight, and form factor (e.g., LIDAR~\cite{OUATTARA202295}). 
In contrast, others require power-hungry radio~\cite{9756977_Niculescu} (up to 100s of \SI{}{\milli\watt}) and additional ad-hoc infrastructure~\cite{10.1145/3498361.3538936}, such as ultra-wideband (UWB) anchors and WiFi routers.
For these reasons, many State-of-the-Art (SoA) nano-drone systems~\cite{BonatoD2D,LiD2D,moldagalieva2023virtual}, including ours, address this problem only by employing cheap vision sensors, e.g., gray-scale low-resolution cameras, and CNNs running aboard nano-drones.
Therefore, \textbf{our contribution} results in a novel vision-based fully convolutional neural network (FCNN), tailored on the ultra-limited resources aboard a commercially-available Crazyflie nano-drone, extended with the Greenwaves Technologies (GWT) GAP8 multi-core System-on-Chip (SoC).
Our FCNN predicts three 20$\times$20 pixel output maps, starting from a gray-scale input image of 160$\times$160 pixels.
Two of the three maps are used to solve the relative pose estimation task, i.e., the 2D position of the drone in the $u,v$ image space and the depth map ($d$).
The third map predicts the per-pixel probability of the target drone's LED state (i.e., on/off), which is a strongly correlated task w.r.t. the pose estimation but outside the scope of this work.

In addition to the FCNN design, we contribute with \textit{i}) full vertical deployment of our FCNN, i.e., from the Python design/training down to the C code execution on the nano-drone's SoC; \textit{ii}) a thorough comparison with three different SoA systems; \textit{iii}) a detailed assessment of our FCNN running onboard the nano-drone, i.e., inference rate and power consumption; \textit{iv}) an in-field evaluation of our closed-loop system regarding endurance and generalization capability.
On a $\sim$\SI{30}{\kilo\nothing} real-world testing images, our FCNN outperforms three SoA models~\cite{BonatoD2D,LiD2D,moldagalieva2023virtual} designed for the same task with the same nano-drone (i.e., output $u,v,$ and $d$).

On average, on our testing set, over the three outputs, our FCNN achieves an $R^2$ score of 0.48 while~\cite{BonatoD2D} scores 0.3,~\cite{LiD2D} obtains -0.57, ~\cite{moldagalieva2023virtual} achieves -0.05.
In terms of onboard inference rate, on the GAP8 SoC, we achieve a real-time performance of \SI{39}{frame/\second}, while \cite{BonatoD2D} achieves \SI{48}{frame/\second}, \cite{LiD2D} achieves $\sim$\SI{5}{frame/\second}, and \cite{moldagalieva2023virtual} achieves $\sim$\SI{5}{frame/\second}.
Finally, when our system is deployed in the field, it shows remarkable performances: \textit{i}) continuous tracking of the target nano-drone for the entire duration of its battery ($\sim$\SI{4}{\minute}); \textit{ii}) generalization capabilities with $\sim$\SI{1}{\minute} uninterrupted flights in three different \textit{never-seen-before} environments; \textit{iii}) and a reduction of the tracking error of 37\%, 52\%, and 23\% on the x, y, and z coordinates respectively, compared to~\cite{BonatoD2D}.

\section{Related work} \label{sec:related_work}

Relative pose estimation between drones can leverage various types of sensors.
Although, given the strict constraints of nano-drones, i.e., payload, power envelope, and size, not all available technologies are affordable, e.g., GPS~\cite{article_Tahar}, LIDAR~\cite{OUATTARA202295}, etc.
GPS-based solutions give their best performance outdoors, while in indoor environments, they estimate the position with a \SI{6}{\meter}-\SI{10}{\meter} error~\cite{Farid}.
Another technology for 3D localization relies on infrared-based systems. 
As an example, the solution proposed in~\cite{article_Roberts} comes with the crucial disadvantage of adding more than \SI{120}{\gram} onboard and thus is unusable on our nano-drone.

Radio-based solutions employing UWB~\cite{pourjabar2023land_uwb, 8481661_Strohmeier_uwb, 9756977_Niculescu} and WiFi~\cite{10.1145/3498361.3538936} can provide accurate pose estimation (sub-\SI{10}{\centi\meter}) and they can be deployed onboard nano-drones~\cite{pourjabar2023land_uwb, 9756977_Niculescu}. 
However, they come with two big disadvantages.
On the one side, they require ad-hoc infrastructure such as UWB anchors~\cite{pourjabar2023land_uwb, 8481661_Strohmeier_uwb}, WiFi routers~\cite{10.1145/3498361.3538936}, etc., that is not always affordable/deployable.
Conversely, UWB-based localization systems need power-hungry devices mounted onboard. 
In~\cite{pourjabar2023land_uwb}, the power consumption for the UWB module aboard the nano-drone peaks at \SI{342}{\milli\watt}, i.e., 5-10$\times$ the power envelope for the onboard computation.
Vision-based systems, instead, use cameras that are available in lightweight and reduced form factors fitting the payload and size of our nano-drone~\cite{BonatoD2D}.
Furthermore, they can operate in ultra-low-power budgets, e.g., sub-\SI{10}{\milli\watt} without any need for ad-hoc infrastructure, e.g., UWB anchors.


CNNs offer a viable solution for drone pose estimation tasks~\cite{BonatoD2D} since they can meet the real-time constraint required for a drone tracking application.
In~\cite{BonatoD2D}, the authors propose a vision-based neural network that solves position estimation as a regression problem.
This network uses as input a 160$\times$96 gray-scale camera frame and produces four scalar variables representing the relative position of the peer drone expressed in $x$, $y$, $z$, and $\phi$, i.e., 3D position in space and the relative yaw angle.
This approach can only work with one nano-drone as a target since the network estimates only one position, therefore unsuitable for swarm missions.
In contrast, our FCNN can tackle multi-drone pose estimation since the output is not bounded a priori to estimate the position of only one drone.
Furthermore, the output of the CNN proposed in~\cite{BonatoD2D} is produced having as a receptive field the entire image possibly failing to capture local details~\cite{article_Gao} that are crucial for this application. 
The nano-drone, in fact, can be as small as 0.06\% of the entire image~\cite{BonatoD2D}.

Works in~\cite{LiD2D, moldagalieva2023virtual} propose a network based on YOLOv3~\cite{Redmon2018YOLOv3AI} that, given an input image produces two maps of 28$\times$40 pixels each.
The first map represents the position in the image space, and the second map estimates the distance.
These approaches are trained with 900 and \SI{50}{\kilo\nothing} simulated images.
The fine-tune is then performed with 192 and $\sim$\SI{8}{\kilo\nothing} real-world images for \cite{LiD2D} and \cite{moldagalieva2023virtual}, respectively.
The networks are then tested with 48 and 250 images, respectively, of the same domain used to perform the fine-tuning.
The approaches proposed by~\cite{LiD2D, moldagalieva2023virtual} need \SI{78.7}{\mega\nothing} multiply-and-accumulate (MAC) operations, which is more than 8.3$\times$ the MAC required by our FCNN.
Finally, assuming the same efficiency achieved with our FCNN, i.e., 2.2 MAC/cycle, these networks would run on the GAP8 SoC at a maximum of \SI{4.6}{frame/\second}, insufficient for effective tracking.
Our work proposes a solution based on FCNN suitable for the deployment on computationally constrained devices such as the GAP8 SoC, still achieving SoA performance in terms of regression performance and throughput, i.e., \SI{39}{frame/\second}.
\section{System design} \label{sec:methodology}

\begin{figure}[t]
  \includegraphics[width=\columnwidth]{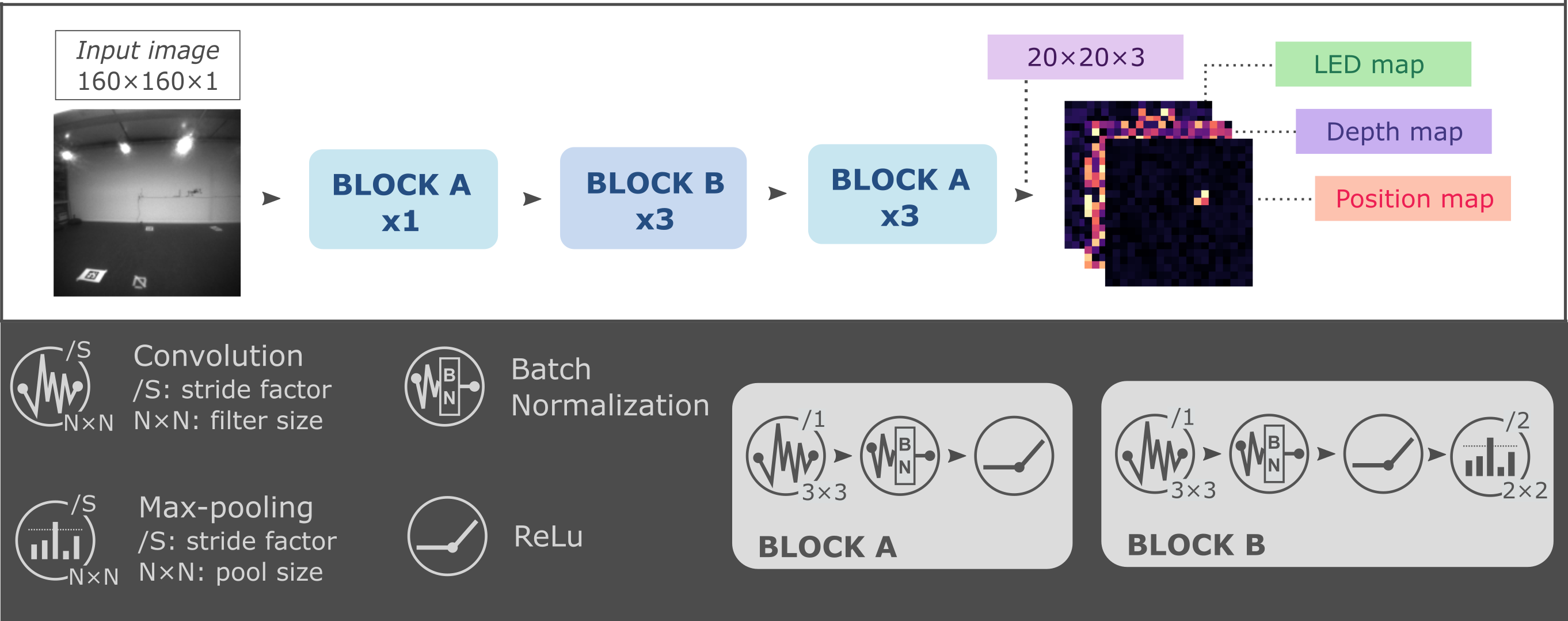}
  \caption{Our fully convolutional neural network feed with grayscale 160$\times$160 images and producing three 20$\times$20 output maps.}
  \label{fig:fully_conv_graphic}
\end{figure}

\paragraph*{Robotic platform}
The nano-drone employed in this work is a \SI{27}{\gram} Bitcraze Crazyflie 2.1, featuring an STM32 microcontroller unit that runs the low-level flight controller and the state estimation task.
The nano-drone is extended with a \SI{4.4}{\gram} AI-deck board, which provides a monocular QVGA grayscale camera (HIMAX HM01B0), a GWT GAP8 SoC, and 8/\SI{64}{\mega\byte} off-chip DRAM/Flash.
The STM32 and the GAP8 can communicate via a UART bidirectional interface.
To increase the drone's stability, we employ a second extension board called Flow-deck, which provides height measurement with a Time-of-Flight (ToF) sensor and $\Delta x$ and $\Delta y$ displacement with a down-looking optical-flow camera.

The GAP8 SoC is designed with two power domains. 
The former, called fabric controller (FC), has one core in charge of data and heavy computation transfer to the latter, the cluster (CL), which leverages eight parallel cores to perform the execution of computationally intensive kernels.
The on-chip memory hierarchy is organized into two levels: a 1-cycle latency \SI{64}{\kilo\byte} L1 memory within the CL and a slower \SI{512}{\kilo\byte} L2 memory.
The SoC provides two direct memory access (DMA) engines that enable efficient transfers between memories and peripherals. 
The GAP8 lacks hardware floating-point units, forcing the adoption of integer-quantized arithmetic to avoid expensive soft-float computations.

\paragraph*{Neural network model}
We tackle the drone pose estimation problem with the FCNN architecture depicted in Figure~\ref{fig:fully_conv_graphic}. 
Given a 160$\times$160 grayscale image, it outputs three 20$\times$20 maps: the LED map expresses the state, on or off, of an LED on the target drone, the depth map represents the distance between the observer drone and the target one, while the position map reports the probability of having the target drone in each pixel.
From the first map, we extract the $(u, v)$ drone image-space coordinates by calculating the barycenter of the activations.
Drone depth is extracted as the weighted average of the depth map, using values of the position map (rescaled such that they sum to 1) as weights.  
The same approach extracts a scalar value for the LED state (probability that the LED is on) from the LED map.

\paragraph*{Datasets}
Training and testing datasets\endnote{\href{https://github.com/idsia-robotics/drone2drone_dataset}{https://github.com/idsia-robotics/drone2drone\_dataset}} have been recorded in a 10$\times$10$\times$\SI{2.6}{\meter} room equipped with an 18-camera OptiTrack motion capture system (mocap).
We recorded 72 flights of $\sim$\SI{210}{\second}, each equally split between the training and the testing sets.
Considering an average acquisition rate of \SI{4}{frame/\second}, we record $\sim$\SI{60}{\kilo\nothing} samples, where the testing and the training set count $\sim$\SI{30}{\kilo\nothing} samples each.
Every sample comprises a 160$\times$160 pixels camera image, the 3D relative pose between the two drones, and the LED state, i.e., on or off, of the target drone.
To train our FCNN, we derive the ground-truth depth map and position map from the recorded 3D pose, and the LED map is obtained from the binary LED state recordings. 
All the ground truth maps (160x160 pixels) are created starting from a map filled with zeros and placing a circle of radius $r=4$ pixels centered in the position ground truth that has a decreasing value from the maximum to 0 with a soft edge transition. 
The maximum value depends on the type of map in fact, the LED map has a maximum value of 1 when the LED is on or 0 when the LED is off, while the position map has a maximum value always equal to 1.
Finally, the maximum value of the depth map is equal to the distance between the observer and the target drone.

\paragraph*{Deployment}
Since the GAP8 SoC does not feature a floating point unit, to avoid the expensive overheads of emulating floating points with integer arithmetic, we deploy our network in the int8 data type using the QuantLib\endnote{\href{https://github.com/pulp-platform/quantlab}{https://github.com/pulp-platform/quantlab}} open-source tool.
Furthermore, we automatically generate C code for the target processor, the GAP8, using DORY~\cite{burrello2020dory}, a tool that relies on the PULP-NN kernels~\cite{garofalo2019pulp}. 
DORY creates the call to the DMA in order to move the tensors from the L2 memory to the L1 memory, which allows the optimized execution of the computations. 
DORY is not limited to L2-L1 transfers and can exploit the complete hierarchy of memories available on the AI-deck, which includes L1, L2, DRAM, and flash memories.
Our network is designed in such a way that all weights, runtime code, and images (double buffered) fit in the L2 memory, i.e., they are under \SI{512}{\kilo\byte}; this prevents heavy overheads for DRAM transfer.
\section{Results}\label{sec:results}

\subsection{Regression performance}\label{sec:Regression_performance}

We evaluate the performance of our model on the testing set by measuring separately for each of the three outputs ($u$, $v$, and $d$): the coefficient of determination metric ($R^2$) and the Pearson correlation coefficient w.r.t. the ground truth.  
The former is a standard metric that measures regression performance in a normalized range [$-\infty$, 1] and reaches 1.0 for a perfect regressor.
A regressor that always predicts the average of the testing set scores 0 if measured with the $R^2$ metric.
The latter metric, the Pearson correlation coefficient, captures the linear correlation between predictions and the ground truths and is unaffected by additive and multiplicative biases.
Table~\ref{tab:comparison_dataset} reports the performance in terms of $R^2$ and the Pearson correlation coefficient of our FCNN compared with the following SoA approaches: a) Li et al.~\cite{LiD2D}; b) Moldagalieva et al.~\cite{moldagalieva2023virtual}; c) Bonato et al.~\cite{BonatoD2D}.

Approaches a) and b) are tested in two configurations each: using the pre-trained networks provided by the authors\endnote{\href{https://github.com/shushuai3/deepMulti-robot}{https://github.com/shushuai3/deepMulti-robot}}\endnote{\href{https://tubcloud.tu-berlin.de/s/Sa5rN5JK7poGawr}{https://tubcloud.tu-berlin.de/s/Sa5rN5JK7poGawr}{ref}}; and using the training approach described by the authors, fine-tuned on our testing environment following the procedure defined in the respective papers~\cite{LiD2D,moldagalieva2023virtual}.
In fact,~\cite{LiD2D,moldagalieva2023virtual} use images from their testing environment to fine-tune their networks.
More specifically, approach a) is first trained on the 800 images training set provided by the authors and then fine-tuned using 192 images from our testing environment; for b) we first train the network with the \SI{50}{\kilo\nothing} images provided by the authors, and then we fine-tuned it with more than 8216 from our testing environment.

Approach c) is a regression model, trained from scratch on our training set, since the pre-trained model and the original dataset are not available, that provides outputs as coordinates in 3D space; to compare them to ours, the outputs in the 3D space are projected back in the $u$, $v$, and $d$ coordinates according to the camera HIMAX camera matrix.
We observe that, on both metrics, our model significantly outperforms all the competing approaches on the $u$ and $v$ variables and performs on par with approach c) on $d$. 
It is worth noting that approaches a) and b) are trained and fine-tuned on different (and smaller) datasets than c) and ours.

\begin{figure}[t]
  \includegraphics[width=\columnwidth]{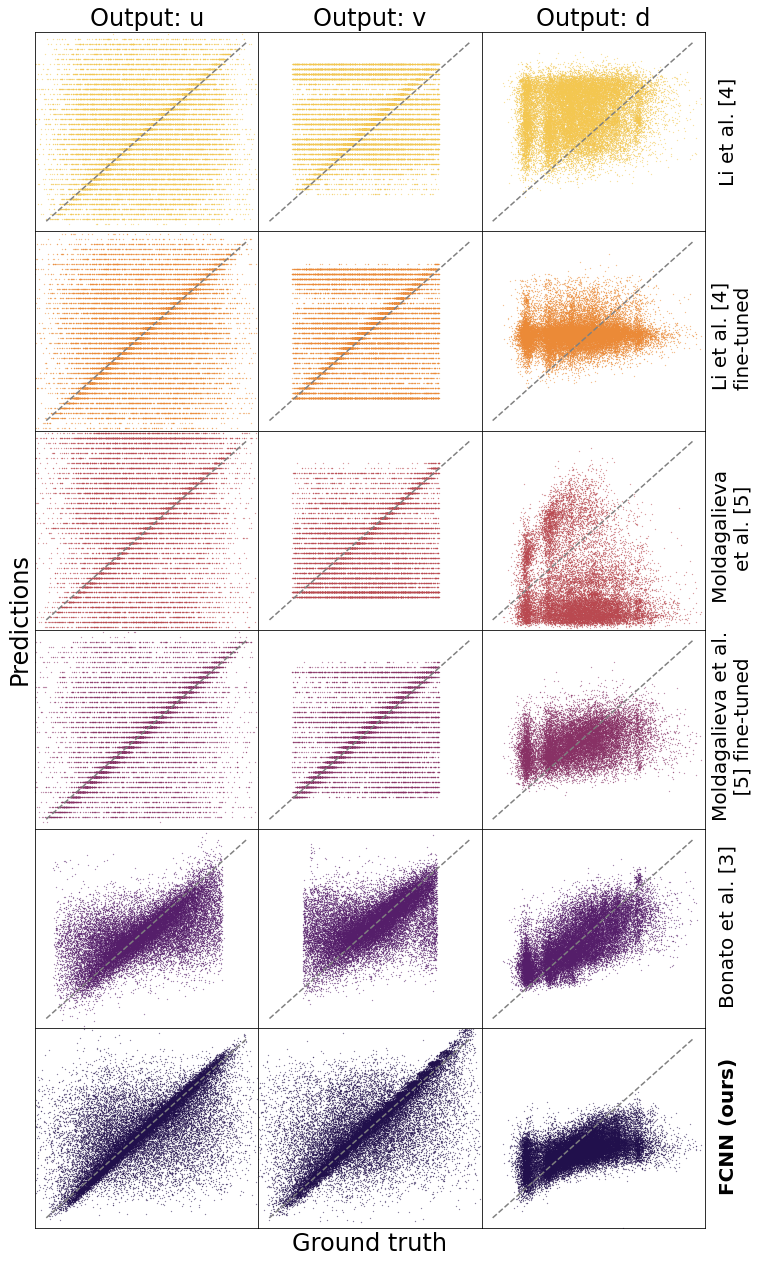}
  \caption{Predictions vs. ground truths for each model (rows), for different outputs (cols). The dashed lines represent a perfect predictor. Some scatter plots are clipped on the output variable $v$ due to the resolution of the input image ranging from 96 to 160 pixels.}
  \label{fig:scatterplot}
\end{figure}

Figure~\ref{fig:scatterplot} compares the predictions vs. ground truths for all the outputs (columns) for each model (rows) with a single dot in a plot representing one testing sample.
The different clipping in the output position $v$ is due to the various input resolutions of each network ranging between 96 and 160 pixels, depending on the model, as described in the previous section.
It's worth noting that the clipping on the output $v$ also reduces the ground truth range on the output $u$ variable in the case of Bonato et al.~\cite{BonatoD2D}.
The scatter plots highlight that our FCNN results in the best-performing network on $u$ and $v$. 
In fact, the distributions of the test samples are the closest to the diagonals (dashed lines), with each diagonal representing a perfect predictor.
Even if our FCNN produces a position map discretized as 20$\times$20 pixels, $u$, and $v$ coordinates are extracted as the barycenter of such map, which yields continuous values.
This is not the case for Li et al. and Moldagalieva et al., who provide discretized values (horizontal lines) since the prediction is computed by selecting the column and row with the maximum value in the output map, i.e., integer values.

\begin{figure*}[t]
  \includegraphics[width=\textwidth]{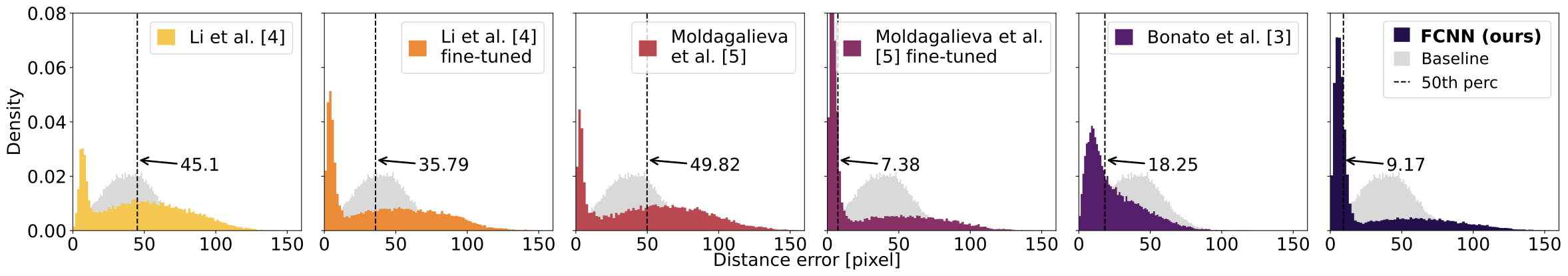}
  \caption{Distribution of image-space distance error between $(u,v)$ predictions and ground truths. All the networks are trained and tested on our dataset.}
  \label{fig:histograms_models}
\end{figure*}

Figure~\ref{fig:histograms_models} reports the distribution of image-space distances between the predicted $(u,v)$ point and the ground truth.
As a lower bound, all graphs report in the background the performance of the dummy predictor baseline (in gray) that always returns the center of the image.
The median error (vertical dashed line) of our approach is $\sim$9 pixels, halving the value achieved by model c); on the other hand, the fine-tuned variant of b) yields a lower median error (7.38 pixels).
This is likely since this model relies on an \texttt{argmax} operation to obtain the output coordinates from the activation map in the last layer, compared to our barycenter approach.
The former is an aggressive approach that provides precise predictions for most samples but yields significant errors for ambiguous cases since it commits to the most likely output; the latter is more conservative, sacrificing accuracy on easy samples to reduce errors on challenging images.
The difference between our FCNN and the fine-tuned version of model b) is visible in Figure~\ref{fig:scatterplot}.
Furthermore, considering also the computational requirements, a) and b) come at the disadvantage of being 8.3 times more computationally expensive than our method and, as such, are unsuitable for applications with high-throughput requirements, such as the tracking task we describe below.

\begin{table}[t]
    \begin{center}
      \caption{Test set regression performance.}
      \label{tab:comparison_dataset}
      \resizebox{\columnwidth}{!}{
      \renewcommand{\arraystretch}{1.25}
      \begin{tabular}{lcccccc} 
        \toprule
        \textbf{Network}  & \multicolumn{3}{c}{\textbf{R2 score [\%]}} & \multicolumn{3}{c}{\textbf{Pearson [\%]}} \\
        \cmidrule(lr){2-4} \cmidrule(lr){5-7}
         & $u$ & $v$ & $d$ & $u$ & $v$ & $d$ \\
        \midrule
        
        \textbf{Li et al.~\cite{LiD2D}}  & -88 & -68 & -174 & 18 & 17 & 16\\
        \textbf{Li et al.~\cite{LiD2D} fine-tuned} & -75 & -66 & -29 & 26 & 23 & 2\\
        \textbf{Moldagalieva et al.~\cite{moldagalieva2023virtual}}  & -161 & -94 & -368 & 18 & 24 & -16\\
        \textbf{Moldagalieva et al.~\cite{moldagalieva2023virtual} fine-tuned} & -12 & -8 & 4 & 50 & 46 & 32\\
        \textbf{Bonato et al.~\cite{BonatoD2D}}  & 32 & 18 & 42 & 58 & 47 & \textbf{66}\\
        \textbf{FCNN}  & \textbf{47} & \textbf{55} & \textbf{42} & \textbf{75} & \textbf{75} & 65 \\
        \bottomrule
      \end{tabular}
      }
    \end{center}
  \end{table}

\subsection{Onboard performance assessment}

\begin{figure}[t]
  \includegraphics[width=\columnwidth]{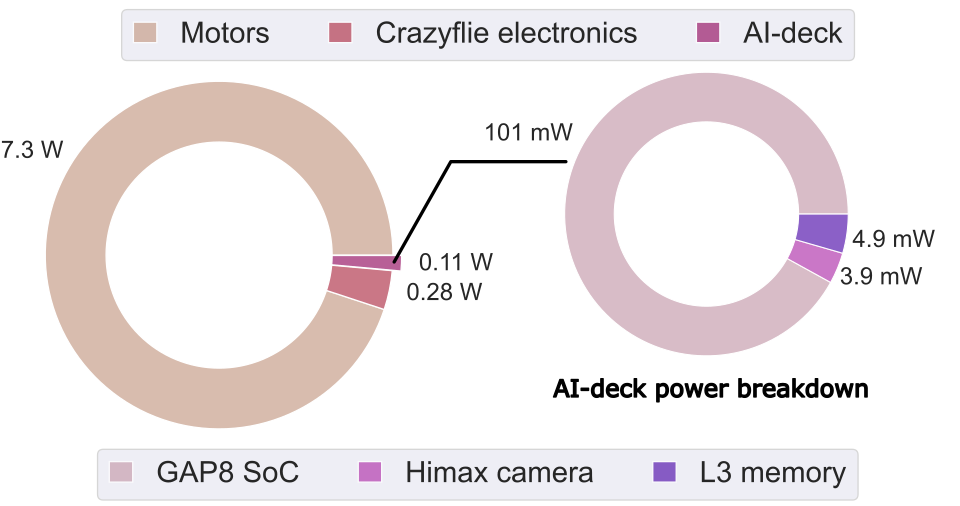}
  \caption{System power breakdown while running the our approach in the \emph{maximum performance} configuration.}
  \label{fig:power_breakdown}
\end{figure}

We evaluate the performance of our FCNN on the GAP8 SoC for what concerns the power consumption and the inference rate in two conditions: \emph{minimum power} (VDD@\SI{1.0}{\volt}  FC@\SI{25}{\mega\hertz} CL@\SI{25}{\mega\hertz}) and \emph{maximum performance} (VDD@\SI{1.2}{\volt}  FC@\SI{250}{\mega\hertz} CL@\SI{175}{\mega\hertz}). 
While the first is useful when the drone acts as a smart sensor, as reported in~\cite{BonatoD2D}, the latter is crucial when the drone flies at high speed, maximizing the onboard inference rate, which feeds the control loops.
In the minimum power configuration, we can process \SI{5.7}{frame/\second} with a power consumption of \SI{10.7}{\milli\watt}.
In the maximum performance configuration, we achieved up to \SI{39}{frame/\second} inference rate -- requiring almost \SI{4.4}{\mega\nothing} cycles per frame on the GAP8 SoC -- with a power consumption up to \SI{100.8}{\milli\watt}.
The total power requirement, including the camera acquisitions, memory transfers, and the GAP8 SoC processing, grows to \SI{109.6}{\milli\watt}.
This power envelope accounts only for 1.43\% of the total power consumption when also considering the drone's electronics and motors, as shown in Figure~\ref{fig:power_breakdown}.

\subsection{LED state classification}
\begin{figure}[t]
    \begin{center}
        \includegraphics[width=0.6\columnwidth]{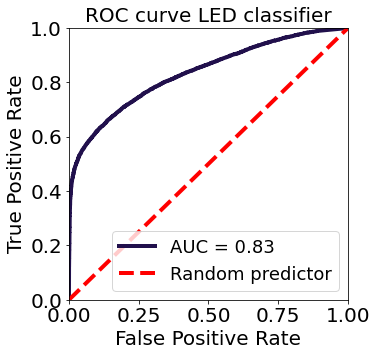}
    \end{center}
  \caption{LED prediction ROC curve for our FCNN.}
  \label{fig:roc_curve}
\end{figure}

We use the third map produced as output by our network to perform the LED classification task as reported in Section~\ref{sec:methodology}.  
We quantify the binary classification performance of the LED state with the Area Under the ROC Curve (AUC) metric.
Figure~\ref{fig:roc_curve} reports the ROC curve for our LED classification task on our testing set, highlighting an AUC score of 0.83.  
A more detailed analysis shows that when the target drone flies at the same or lower height as the observer, the AUC exceeds 0.90. 
This indicates excellent classification performance and enables applications with LEDs used as low-bandwidth communication channels.
In contrast, when the target drone flies higher than the observer one, the LEDs -- placed on the top side of the drone frame -- are invisible, and classification becomes impossible in many frames (AUC $<$ 0.70).

\subsection{In-field evalutation}

\textbf{SoA comparison.}
We perform extensive in-field tests of our FCNN compared with the SoA approach of Bonato et al.~\cite{BonatoD2D}. 
The networks based on~\cite{LiD2D} and~\cite{moldagalieva2023virtual} are not tested in-field due to their limited frame rate (i.e., \SI{5.2}{frame/\second}).
The setup consists of two drones: a target drone and an observer one.
The target drone flies a pre-defined 3D spiral path at a constant speed ($\sim$\SI{0.21}{\meter/\second}), as defined in~\cite{BonatoD2D}.
The observer tracks and follows the target employing $u$, $v$, and $d$ outputs of our FCNN.
We assume the observer drone's $x$ axis is aligned with the world's $x$ axis. 
The desired position of the observer drone is in front of the target one, keeping a constant distance of \SI{0.8}{\meter} (x-axis).

\begin{figure}[t]
  \includegraphics[width=\columnwidth]{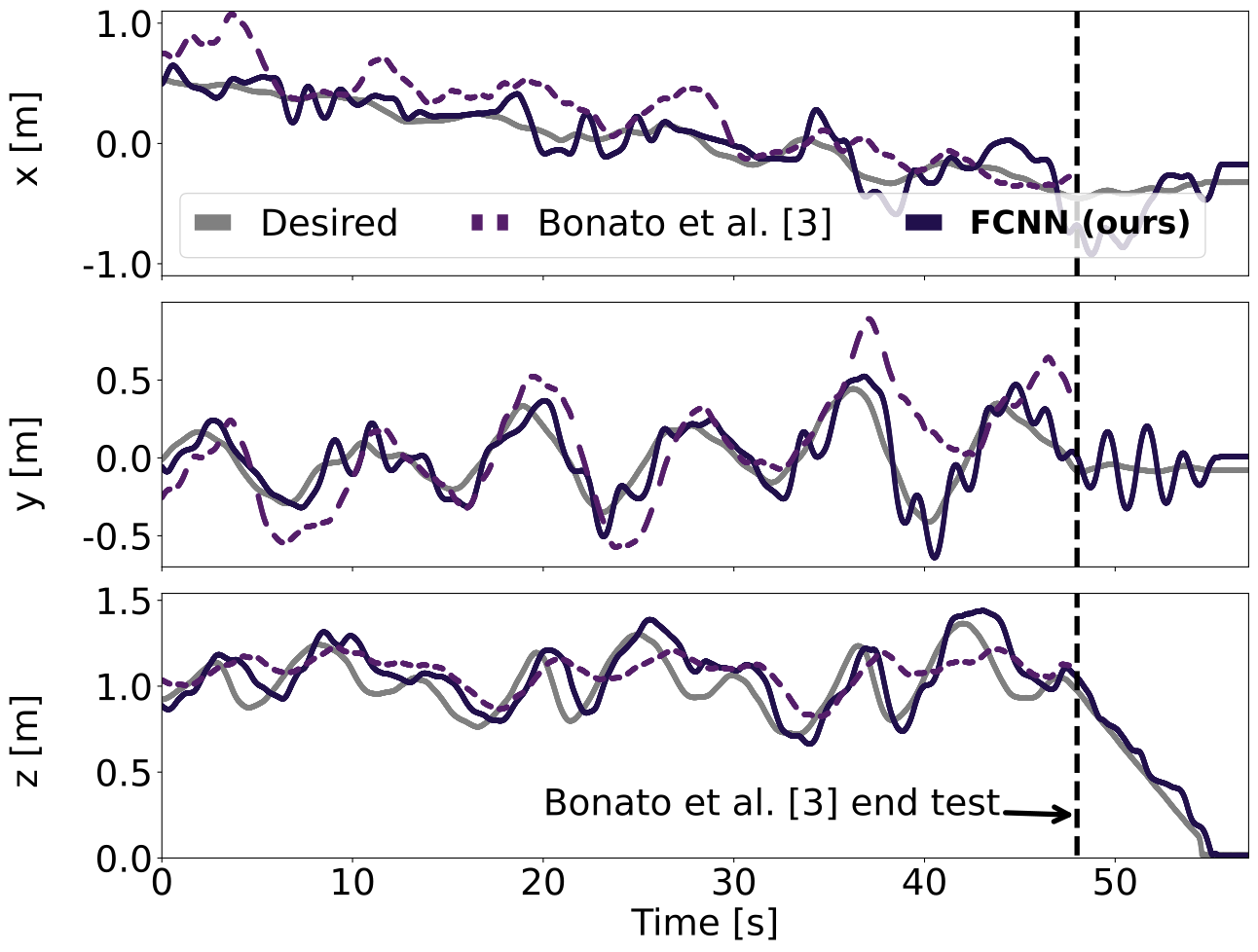}
  \caption{Trajectory comparison of the observer drone controlled with two approaches vs the desired trajectory.}
  \label{fig:comparison_infield}
\end{figure}

Figure~\ref{fig:comparison_infield} reports the in-field performance of the two tested approaches, i.e., Bonato et al. (dashed violet line) and our FCNN (continuous violet line), w.r.t. the desired position (gray line) over a time-frame of $\sim$\SI{60}{\second}.
The performance of the drone controlled with our FCNN is remarkably better, if compared to Bonato et al., for the whole duration of the experiment, achieving an average position tracking error comparable to the diameter of our Crazyflie nano-drone, i.e., the error is lower than \SI{10}{\centi\meter} on each axis. 
Furthermore, as displayed in Figure~\ref{fig:comparison_infield}, after the vertical dashed line, we achieved precise tracking of the target drone while landing. 
The observer drone lands with a position error of $\sim$\SI{15}{\centi\meter} with respect to the desired landing position.
The in-field tracking error of this test is reported in \ref{tab:infield_results} for the two networks with the postfix text ``v 0.21''. 
The notation ``v 0.21'' represents the average speed of the target during the path, namely \SI{0.21}{\meter\per\second}. 
In this setting, our approach reduces the average position tracking error by 37\%, 52\%, and 23\%, respectively, on x, y, and z if compared to the approach proposed by Bonato et al.~\cite{BonatoD2D}.
All the reported metrics have been computed in the time window where both systems were working, i.e., before the vertical dashed line in Figure~\ref{fig:comparison_infield}.

\textbf{Increasing target velocities. }
The system with our FCNN has been tested five times on the same 3D spiral path described previously, per each configuration of speed, in particular for $v=\SI{0.21}{\meter\per\second}$, $v=\SI{0.34}{\meter\per\second}$, and $v=\SI{0.59}{\meter\per\second}$.
In Table~\ref{tab:infield_results}, we report the average tracking error on each world-axis and the aggregate tracking error ($|p-p_d|$) based on the distance between the observer pose ($p$) and the desired pose ($p_d$) in the 3D space.
In all the experiments, our FCNN running onboard the observer drone successfully tracks the target drone for the complete path.
Our observer drone can track a target drone in front of it with a maximum speed of \SI{0.61}{\meter\second}, which is $2.8\times$ higher than the speed of the target drone in Bonato et al.~\cite{BonatoD2D}.
This is possible thanks to the higher throughput (\SI{39}{\hertz}) and the improvement in the position estimation of the target drone, as reported in Table~\ref{tab:comparison_dataset}.

\begin{table}[t]
    \begin{center}
      \caption{In-field tracking performance (average over five runs) with changing average speed}
      \label{tab:infield_results}
      \resizebox{\columnwidth}{!}{
      \renewcommand{\arraystretch}{1.25}
      \begin{tabular}{lcccccc} 
        \toprule
        \textbf{Configuration}  & completed runs & \multicolumn{5}{c}{\textbf{Avg tracking error [m]}} \\
        \cmidrule(lr){3-7}
         &  & $x$ & $y$ & $z$  & $|p-p_{d}|$ & $\sigma |p-p_d|$\\
        \midrule
        \begin{tabular}{@{}c@{}} \textbf{Bonato et al.~\cite{BonatoD2D}}\\\textbf{v 0.21}\end{tabular} & not reported & 0.16 & 0.21 & 0.13 & not rep. & not rep.\\
        \midrule
        \textbf{FCNN (ours) v 0.21} & 5/5 & 0.09 & 0.10 & 0.10 & 0.19 & 0.08\\
        \textbf{FCNN (ours) v 0.34}& 5/5 & 0.10 & 0.15 & 0.14 & 0.26 & 0.12\\
        \textbf{FCNN (ours) v 0.59} & 5/5 & 0.12 & 0.31 & 0.16 & 0.41 & 0.15\\
        \bottomrule
      \end{tabular}
      }
    \end{center}
  \end{table}

\begin{figure}[t]
  \includegraphics[width=\columnwidth]{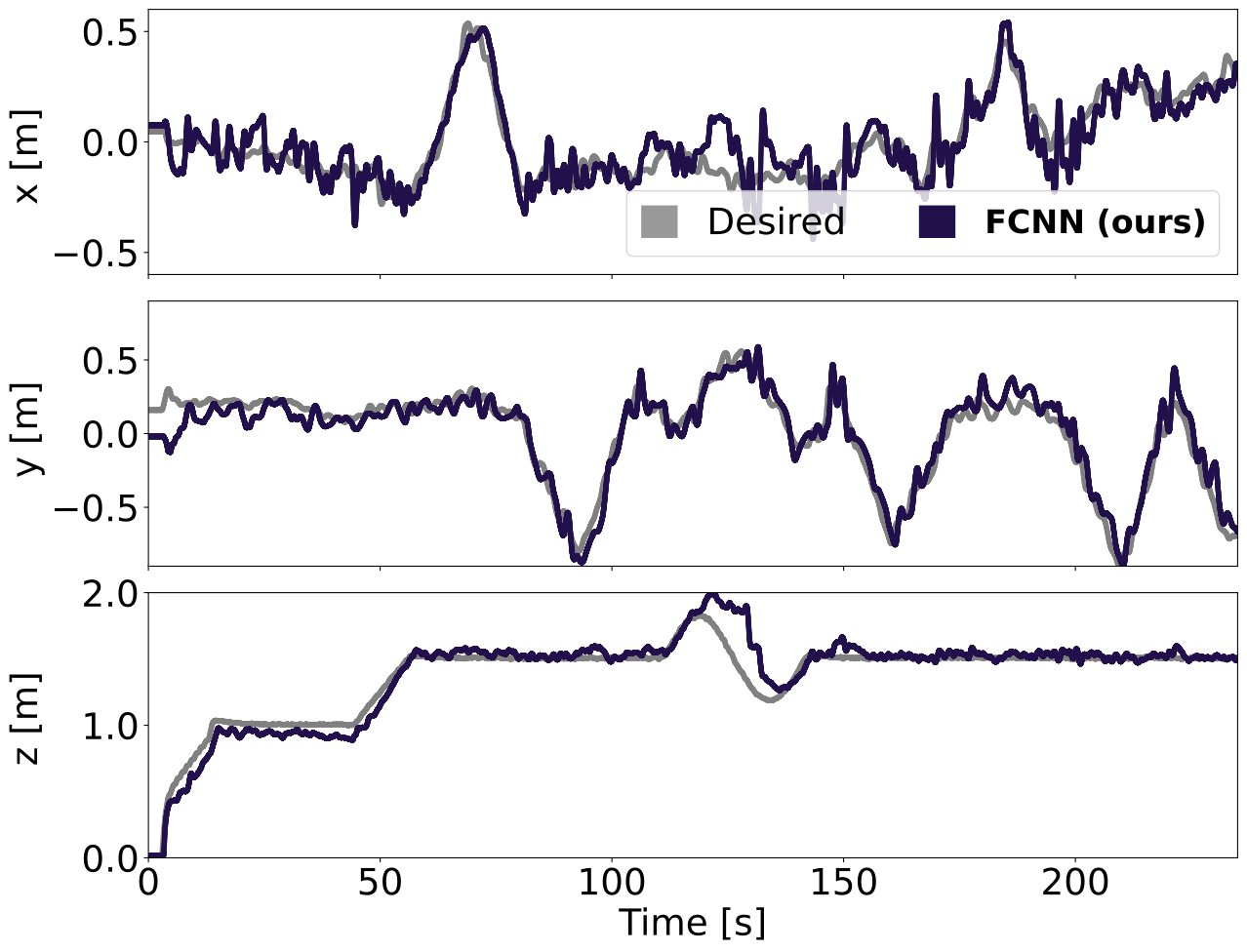}
  \caption{In-field endurance test of our FCNN model for up to \SI{240}{\second}.}
  \label{fig:endurance}
\end{figure}

\begin{figure}[t]
  \includegraphics[width=\columnwidth]{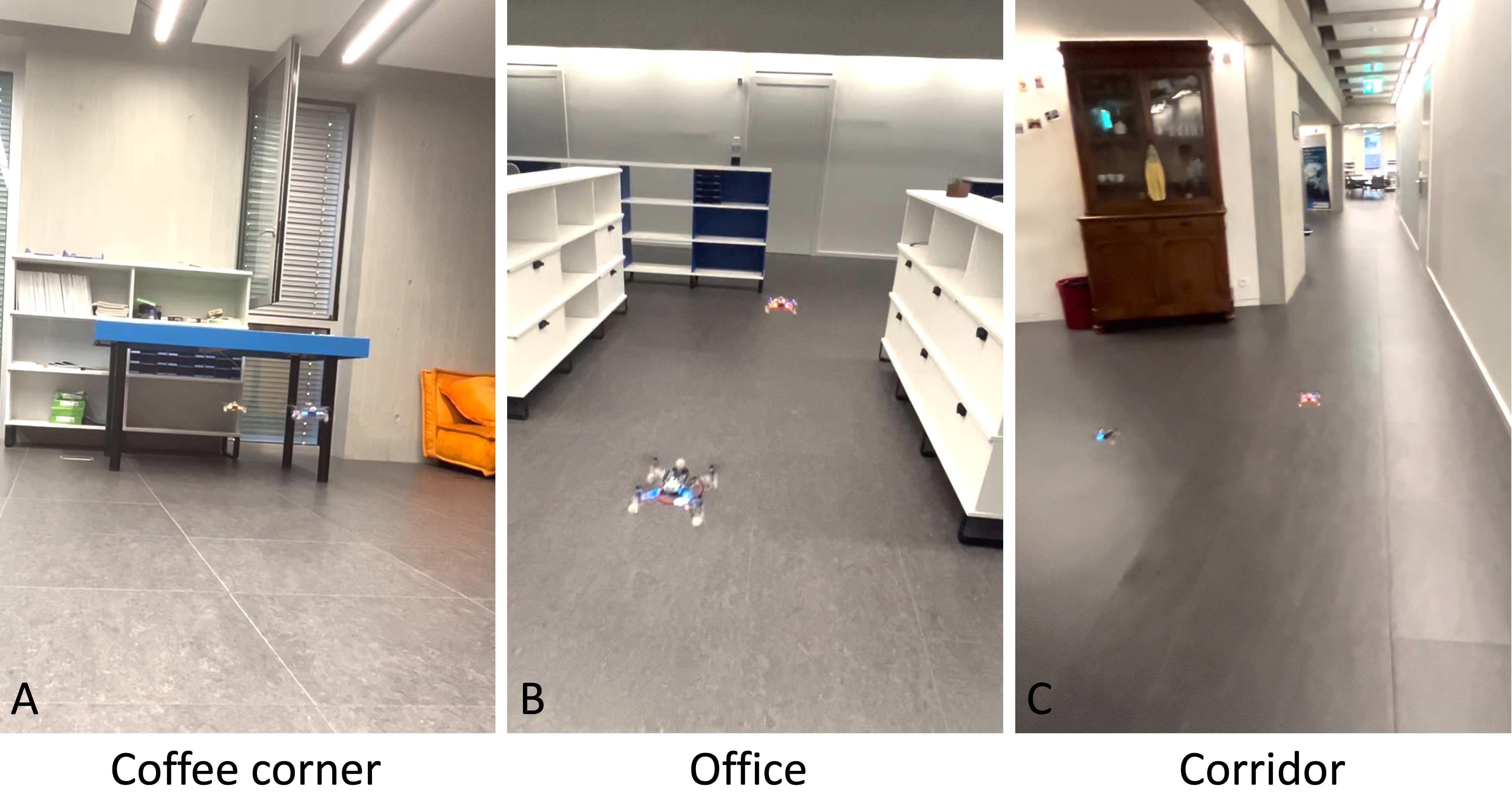}
  \caption{Generalization in three different environments.}
  \label{fig:generalization}
\end{figure}

\textbf{Endurance test.}
In Figure~\ref{fig:endurance}, we report an endurance test running for the entire duration of the nano-drone's battery, i.e., \SI{380}{\milli\ampere\hour}, which lasts for \SI{240}{\second}.  
The target performs linear movements separately for each axis and circles composed of movements on the 3 axes stressing the accuracy of the predictions in the 3 DoF.
The observer drone can track the target for the entire experiment duration, achieving on average (5 runs) a tracking error of \SI{0.08}{\meter}, \SI{0.07}{\meter}, and \SI{0.06}{\meter} for x, y, and z, respectively.

\textbf{Generalization test.}
Furthermore, our system has been tested in three additional environments different from the one in our training set, with several objects never seen before, such as chairs, sofas, and bookcases.
Despite, we can not provide accurate quantitative measurements of the tracking performance due to the lack of a mocap system in our three generalization environments, i.e., coffee corner, office, and Corridor (Figure~\ref{fig:generalization}), we provide a video (see supplementary video material) as a qualitative evaluation of our system in these never-seen-before rooms.
In this generalization test, the target nano-drone is remotely operated through a controller while the observer drone runs the FCNN.
\section{Conclusion} \label{sec:conclusion}

This work addresses the drone-to-drone visual localization task by employing resource-constrained nano-drones.
We propose a novel lightweight FCNN, i.e., 8$\times$ fewer operations than SoA solutions~\cite{LiD2D,moldagalieva2023virtual}, integrated and deployed on a nano-drone extended with a GWT GAP8 SoC.
On a \SI{30}{\kilo\nothing} samples real-world testing dataset, our model marks an $R^2$ score of 0.48 while \cite{BonatoD2D} obtains 0.3, \cite{LiD2D} scores -0.57, and \cite{moldagalieva2023virtual} achieves -0.05. 
Our FCNN reaches an inference rate up to \SI{39}{\hertz} within only \SI{101}{\milli\watt}.
In-field tests demonstrate on average 37\% lower tracking error, compared to~\cite{BonatoD2D}, which, to the best of our knowledge, is the only SoA approach deployable onboard a nano-drone to perform a pose estimation task of another nano-drone in front of it.
Furthermore, with our FCNN, we continuously track a peer nano-drone for the entire nano-drone's battery lifetime, i.e., \SI{4}{\minute}.
Finally, our FCNN shows remarkable generalization capabilities by continuously tracking a target nano-drone even in three never-seen-before environments.


\bibliographystyle{./IEEEtran}
\theendnotes
\bibliography{biblio}

\end{document}